\documentclass{sig-alternate-05-2015}
\usepackage{float}
\usepackage{lipsum}
\usepackage[ruled]{algorithm2e}

\begin{document}




\conferenceinfo{CIKM '16}{Oct 24--28, 2016, Indianapolis, IN, USA}


%
\conferenceinfo{CIKM Cup}{2016, Indianapolis, USA}

\title{Cross Device Matching for Online Advertising with Neural Feature Ensembles : First Place Solution at CIKM Cup 2016} 

%
%
%
%
%

\numberofauthors{3} 
%
\author{
%
%
\alignauthor
Minh C. Phan\\
       \affaddr{Nanyang Technological University}\\
       \email{phan0050@e.ntu.edu.sg}
\alignauthor
Yi Tay\\
       \affaddr{Nanyang Technological University}\\
       \email{ytay017@e.ntu.edu.sg}
\alignauthor
Tuan-Anh Nguyen Pham\\
       \affaddr{Nanyang Technological University}\\
       \email{pham0070@e.ntu.edu.sg}
}


\maketitle
\begin{abstract}
We describe the $1st$ place winning approach for the CIKM Cup 2016 Challenge. In this paper, we provide an approach to reasonably identify same users across multiple devices based on browsing logs. Our approach regards a candidate ranking problem as pairwise classification and utilizes an unsupervised neural feature ensemble approach to learn latent features of users. Combined with traditional hand crafted features, each user pair feature is fed into a supervised classifier in order to perform pairwise classification. Lastly, we propose supervised and unsupervised inference techniques. The source code\footnote{In the case where the repository is not accessible, please contact the authors} for our solution can be found at \textit{http://github.com/vanzytay/cikm\_cup}. 
\end{abstract}

%
%


%
%

%
%
\printccsdesc



\section{Introduction}

Online Advertising is a crucial and essential component of business strategy. For many advertising companies, the ability to serve relevant ads to users is a desirable and attractive prospect. However, it is commonly accepted that users may own or use multiple devices. In order to leverage the rich and sophisticated user profiles learned, it would be ideal to be able to identify the same person across devices. In this paper, we describe the $1st$ place winning approach in the CIKM Cup 2016 Challenge\footnote{https://competitions.codalab.org/competitions/11171}. The problem at hand is intuitive and simple. Given browsing logs of users, generate a list of candidate user pairs that are predicted to be the same person. This can be seen as a ranking problem which we are able to conveniently cast as pairwise classification. In this paper, we describe our approach and findings. 

\section{Problem Formulation}
In this section, we describe the dataset and experimental evaluation metrics.

\subsection{Dataset}
The dataset used in this competition was provided by DCA (Data Centric Alliance). In general, the dataset is generally comprised of user browsing logs with timestamps. Each user click is defined as an event with a timestamp and url. For a certain fraction of urls, meta-data such as title text is provided. All words in the dataset (in urls and title text) are hashed with MD5. This includes url paths (words in url paths have the same hash code if it appears in any meta-data). For supervised learning, the dataset includes $506,136$ training pairs. The following table describes the characteristics of the competition dataset.

 \begin{table}[htbp]
   \centering
     \begin{tabular}{|l|r|}
     \hline
     \# Training Pairs & 506,136 \\
      \# Testing Pairs & 215,307 \\
     \# Events & 66,808,490 \\
     \# Domain Names & 282,613 \\
     \# Tokens (Urls) & 27,398,114 \\
     \# Tokens (Titles) & 8,485,859 \\
     \# Unique Users & 339,405 \\
     \# Unique Websites & 14,148,535 \\
     \# Users in Train & 240,732 \\
     \hline
     \end{tabular}%

   \label{tab:addlabel}%
      \caption{Dataset Characteristics}
 \end{table}%

\subsection{Evaluation Metric}
The goal of the competition is to identify the same users across multiple devices. The evaluation metric used in the leaderboard ranking is the F1 measure. In each stage of the competition, 215,307 ground truths are used in a $50-50$ split for the validation stage and final test phrase. In most cases, the optimal number of pairs submitted at each stage is a value determined by the contestants.

\section{Our Method}
\begin{figure}[!t]
	\begin{center}
		\includegraphics[scale=0.50]{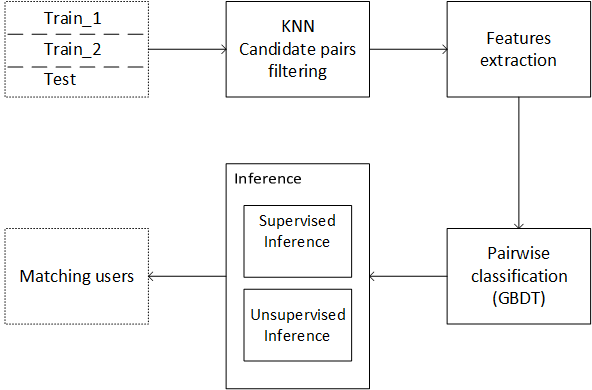}
		
		\caption{\scriptsize{Overall Flow and Architecture}}
		\label{overall}
	\end{center}
\end{figure}

This section describes the framework used in this competition. Firstly, we perform candidate selection (a combination using TF-IDF baselines and neural language models) to select likely candidates. For example, we take $k$ nearest neighbors from vector representations of users as selected candidates. Since this often results in a huge excess of candidate user pairs, we have to perform candidate filtering. We do that via supervised pairwise classification. In general, we regard the problem as a pairwise classification problem. By performing pairwise classification, we use the likelihood scores $f(u_1,u_2) \in [0,1]$ to rank candidate pairs. Finally, we include a supervised and unsupervised inference technique to refine the final selected candidates. Figure \ref{overall} describes the architecture of our approach.

From 240,732 users in training set, we sample 98,000 users as \textit{train 1} which will be used to train the pairwise classifier. An additional 98,000 users, labeled \textit{train 2} are used for training a supervised inference classifier.

\subsection{Candidate Selection}
\begin{figure}[t]
	\begin{center}
		\includegraphics[scale=0.36]{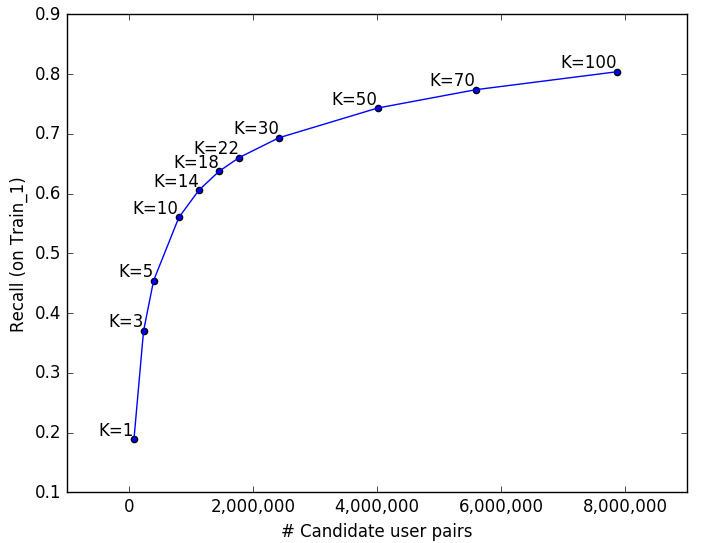}
		
		\caption{\scriptsize{Recall on KNN results}}
		\label{recall_knn}
	\end{center}
\end{figure}

We perform candidate selection via a myriad of user representation techniques. The first obvious representation to use is the TF-IDF vector representation of each user. To construct TF-IDF vectors, we construct $n$ tokens for each level in the url hierarchy, i.e., $a/b/c$ becomes 
$[a,ab,abc]$ For each user, we generate $k$ nearest neighbors and add them to the list of prospective candidates. In addition, we generate prospective candidates using models trained by Neural Language Models. To select an appropriate value for $k$, we studied the recall levels on the development set. Based on Figure \ref{recall_knn} which shows the recall levels with varying $k$, we selected $k=18$ as a trade-off between recall and classification performance.

\subsection{Learning User Features}
This section describes the feature engineering process. We used a combination of hand crafted features along with unsupervised feature learning methods. Our approach heavily relies on unsupervised features instead of manual features. It is good to note that our unsupervised features were dominant in our \textit{feature importance} analysis which can be derived from gradient boosting classifiers.

\subsubsection{Hierarchically Aware Neural Ensemble}
Our approach heavily exploits Neural Language Models\footnote{We tried LDA but found it lacking as compared to modern language modeling approaches} to learn semantic user representations. We use Doc2Vec, an extension of Word2Vec \cite{DBLP:conf/nips/MikolovSCCD13} to learn semantic representations of users.  
Intuitively, we simply treat the sequential click history of a user as words in a sentence. Therefore, a user and his url clicks are analogous to words in a document. The key idea of these models is that they learn based on global concurrence information.  However, unlike words in a document, there is a rich amount of hierarchical information present in url information that can be exploited. Therefore, for each url hierarchical level $h=\{0,1,2,3\}$, we learn a separate Doc2Vec model. In this case, $h=0$ simply means we use simply the domain name. It is good to note that there will be more duplicate tokens in a string of url sequences at lower hierarchical levels. To deal with this, we remove all consecutive duplicate items when training our Doc2Vec model. Furthermore, we also train a \textit{word-level} Doc2Vec model by treating only considering word tokens from urls that include titles. Overall, the output of each Doc2Vec model is a unique semantic vector representation of each user. 

For our implementation, we used the Gensim\footnote{https://radimrehurek.com/gensim/} Package for training our models. We use the default setting for all our models but steadily decrease the learning rate and trained additional models with varying window sizes amongst ($W=\{5,10\}$). We also experimented with the \textit{concat} model that concatenates the vectors at the hidden layer instead of averaging. The dimensionality of each model is $d=300$. We also prune infrequent tokens/urls that appears less than $5$ times in the entire dataset.

\subsubsection{From Neural Models to Feature}
Given that each of our model produces user vectors of length $300$ dimensions, it would be impractical to use these vectors directly as features. Empirically, we found that using these vectors directly seem to worsen performance. Thus, for generating features from these neural models, we use distance measures such as the manhattan, euclidean and cosine distance/similarity between each user pair as features. In addition, we add the order information between two users as well. This is done by taking the k-nearest neighbours of a user A and computing the rank of user B's appearance in that list. This is done vice versa as well.


\subsubsection{Time Features}
To supplement our latent semantic features of users, we included time features. For each user, we build the user's usage pattern by counting the number of urls visited for each hours of the day (hourly pattern), and for each day of a week (weekly pattern). We calculate the absolute difference for the raw counts and \textit{Kullback Lieber} difference for the normalized counts (distributions) as the time features for each user pair. We also extract the number of overlaps between user's logs for each time interval of $\{5,10,60\}$ minutes.

\subsubsection{Feature Importance Test}
\begin{figure}[t]
\begin{center}
\includegraphics[scale=0.2]{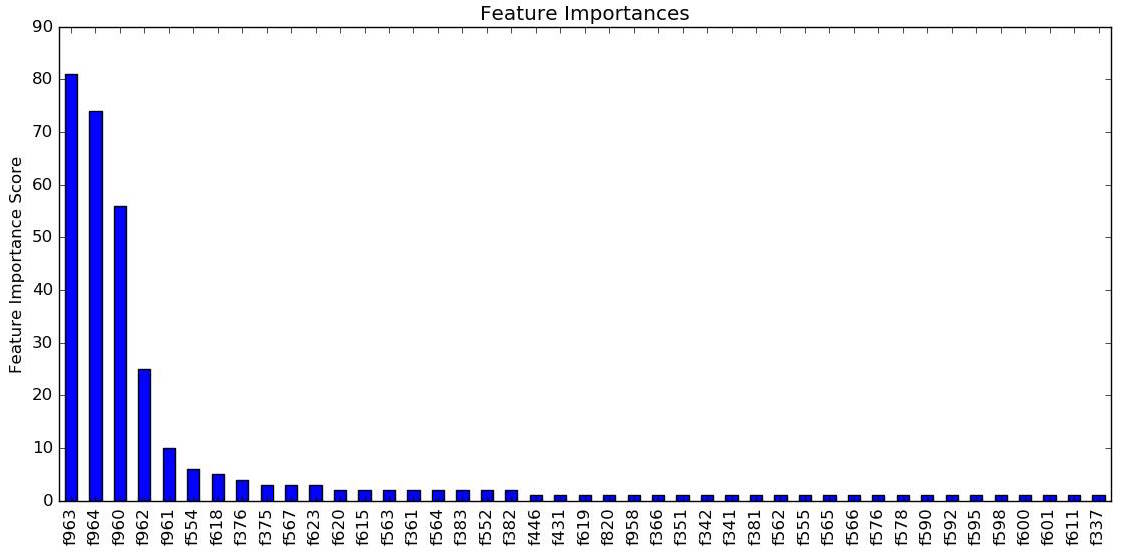}

\caption{\scriptsize{Feature Importance Analysis from XGB}}
\label{featureimportance}
\end{center}
\end{figure}

We generated a feature important test\footnote{Note that these features were generated mid-way for analysis purposes and are not the features from our final model} from XGB. Figure \ref{featureimportance} shows the visualisation of feature importance. Not surprisingly, our neural features are the most important features.  $f963, f964$ and $f960$ are features from the Doc2Vec model (cosine similarity) while $f962$ and $f961$ are TF-IDF features. 

\subsection{Pairwise Classification}
Next, we train a supervised pairwise classifier (XGB1) to predict the likelihood of each user pair being the same user. Given our sampled train users (98,000) we sample positive and negative pairs from knn candidate filtering (with $k=18$) to train the model. Note that we deliberately sample negative samples from the pool of near neighbors instead of random corruption. We found this significantly more effective since it gives our model more discriminative ability since it provides our classifier more challenging negative examples.

\subsubsection{Supervised Classifier}
Our approach mainly used XGBoost \cite{DBLP:conf/kdd/ChenG16}, a gradient boosting technique that is prevalent in the winning solutions of many machine learning competitions. We tried several other classifiers such as Random Forests and the standard Multi-Layer Perceptron (MLP). In general, we found that the MLP did not come close to the performance of tree based ensemble classifiers. Our experiment evaluation section describes and reports on the results for all the algorithms that we tried.

\subsection{Inference Techniques}

In this section, we introduce the inferencing techniques used to increase the performance results. We view the original user linking problem as a clustering problem in the way that each group of same users is associated with a cluster. If there is a link created for two users between cluster A and B, we also make new links for other users across cluster A and B. Specifically, we sort the candidate pairs based on the $XGB1$'s confidence score and alternately taking the sorted pairs to perform the merging. The details are shown in Algorithm~\ref{alg:inf_algo}. In order to control the created clusters's expansion, we propose the following two inference methods. 

\begin{figure}[t]
	\begin{center}
		\includegraphics[scale=0.6]{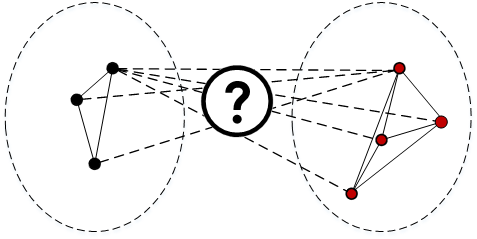}
		
		\caption{\scriptsize{Clustering user group}}
		\label{inference}
	\end{center}
\end{figure}

\begin{algorithm}[t]
	\label{alg:inf_algo}
	\SetKwInOut{Input}{input}
	\SetKwInOut{Output}{output}
	\Input{sorted pairs $\mathbf{S}$, inference method $\mathbf{I}$}
	\Output{extended pairs $\mathbf{E}$}
	
	$\mathbf{E}$=$\emptyset$\;
	Assign each user in $\mathbf{S}$ a unique cluster label\;

	\For{$each pair (u,v) \in \mathbf{S}$}{
		$l_u$ = cluster\_label($u$)\;
		$l_v$ = cluster\_label($v$)\;
		
		\If{$l_u \neq l_v  \wedge \mathbf{I}.cond\_satisfied(l_u, l_v)$}{
				$merge\_clusters(l_u, l_v)$\;
				\For{$i \in cluster(l_u)$}{
					\For{$j \in cluster(l_v)$}{
						$\mathbf{E} \leftarrow (i, j)$\;
					}
			
				}
		}
	}

	\Return $\mathbf{E}$
	\caption{Inference by cluster merging}
\end{algorithm}

\subsubsection{Supervised Inference}
For supervised inference, we only merge two clusters if the average vote of all pairs across 2 clusters is greater than a threshold $\alpha=0.5$. The votes is the confidence scores of 2 users to be merged. In order to have this confidence score, we train another pairwise classifier and refer it as $XGB2$. Note that $XGB2$ is different with  $XGB1$. The training data for $XGB2$ is the extended pairs from Algorithm~\ref{alg:inf_algo} with a $'blind'$ inference that always allow the merging of 2 clusters.

\subsubsection{Unsupervised Inference}
 Our empirical analysis of the training set shows that most of the users are owning 3-5 devices on average. Based on the simple intuition of $(a,b) \wedge (b,c) \rightarrow (a,c)$, we generate new candidate pairs. Naturally, we limit the total size of clusters to be merged to a threshold $\beta=5$ since we regard such inference process \textit{risky}. This is because we are assuming all pairs used in the inference are correct. Our empirical observations on local testing shows that such inferencing may run a risk of decreasing the F1-score. However, when tuned correctly, i.e., selecting the right threshold of candidate pair scores to be included in the inference process, can lead to major increase in performance.

\subsubsection{Final Candidate Selection}
\begin{figure}[t]
\begin{center}
\includegraphics[scale=0.45]{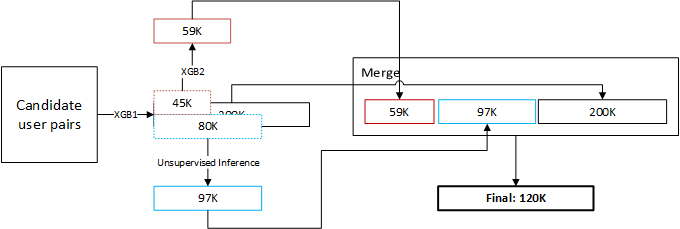}

\caption{\scriptsize{Overview of Final Candidate Selection}}
\label{inference2}
\end{center}
\end{figure}

We use the top 45,000 sorted pairs together with the supervised inference as the input for Algorithm~\ref{alg:inf_algo} and obtain about 59,000 extended pairs. Similarly, using top 80,000 pairs with the unsupervised inference gives us 97,000 extended pairs. We combine the pairs extended from supervised inference, unsupervised inference and sorted pairs from the pairwise classifier (XGB1) by that order. The final pairs for submission will be the first 120,000 unique pairs in the combined list. Figure \ref{inference2} illustrated the selection procedure.

\section{Experimental Evaluation}
This section outlines the experimental evaluation on our local test set as well the competition leaderboard.

\subsection{Local Development Tests}
For our model development, we experimented with several different classifiers. Table \ref{local_eval} shows the results
on our local development set at F1@$120K$ which according to our empirical observations, is an optimal split for the number of pairs to submit. For the sake of completeness, we included both supervised and supervised techniques. 
 \begin{table}[htbp]
   \centering
 
     \begin{tabular}{|c|r|}
     \hline
     \textbf{Method} & \multicolumn{1}{c|}{\textbf{F1@120K}} \\
     \hline
     TF-IDF + KNN  & \multicolumn{1}{c|}{0.065} \\
     Doc2Vec + KNN & \multicolumn{1}{c|}{0.072} \\
     LSTM-RNN & \multicolumn{1}{c|}{18.22} \\
     MLP  & \multicolumn{1}{c|}{25.63} \\
     Random Forest & \multicolumn{1}{c|}{38.32} \\
     XGB   & \multicolumn{1}{c|}{41.60}  \\
     \hline
     \end{tabular}%
       \caption{Evaluation on Local Development Set}
   \label{local_eval}%
 \end{table}%

 Aside from the staple gradient boosting methods, we experimented with deep learning techniques namely the standard Multi-layer Perceptron (MLP) and Recurrent Neural Networks. For MLP, XGB and RF, we trained it when the exact same features described in earlier sections. We also experimented with using user clicks as sequential inputs to a long short term memory recurrent\footnote{This was generally unstable and takes a long training time. Without satisfactory results, we quickly abandoned this approach. However, we note that the final output state might be a useful feature} neural network (LSTM-RNN) as a binary classifier with a final sigmoid layer. 

It is clear that unsupervised techniques such as the TF-IDF + KNN baseline does not perform well. It is good to note that Doc2Vec
performs slightly better than TF-IDF which is expected. Amongst supervised classifiers, our XGB performs the greatest. We quickly abandoned MLP due to time and hardware constraints since we could not afford the computational resources to intensively tune the hyperparameters. Similarly, we did not perform hyperparameter tuning for RF and XGB except for the number of trees. Our final XGB used about $3500$ trees. 
\subsubsection{On Training Time}

The training time for XGB took about $\approx 4$ hours on our machine. MLP was significantly faster at $\approx 30$ minutes using GTX980 GPUs. However, LSTMs, due to needing to process long sequences of user streams took 2 days even with GPUs. Generating prediction scores for users also took up a non-trivial amount of time often amounting to $\approx 1-2$ hours easily.  

\subsection{Final Competition Evaluation}
Finally, we report on the final competition results. In total, we submitted 120,000 pairs for evaluation as a trade-off between precision and recall. 

 \begin{table}[htbp]
   \centering
   
   \setlength{\tabcolsep}{4pt}
     \begin{tabular}{|c|c|c|c|c|}
     \hline
     \textbf{User} & \textbf{Team} & \textbf{F1} & \textbf{Precision} & \textbf{Recall} \\
     \hline
     ytay017 & NTU   & \textbf{0.4204} & 0.3986 & \textbf{0.4445} \\
     ls    & Leavingseason & 0.4167 & 0.3944 & 0.4416 \\
     dremovd & MlTrainings 1 & 0.4120 & \textbf{0.4031} & 0.4213 \\
     bendyna & MlTrainings 2 & 0.4017 & 0.3659 & 0.4452 \\
     namkhantran & -     & 0.3611 & 0.3323 & 0.3954 \\
     \hline
     \end{tabular}%
     \caption{Final Evaluation Results}
   \label{tab:addlabel}%
 \end{table}%

Our final submission consist of a single classifier, XGB with $3500$ trees using the above mentioned features. We applied inferencing techniques mentioned above to generate the final submission file. It is good to note that unsupervised inference increased our final F1 score from $0.4155$ to $0.4204$ which was critical to win the competition.


\section{Conclusion}
We proposed a framework for cross device user matching. Our system involves a novel application of neural language models onto clickstream information. We increased the F1 of the baseline by $751\%$. The major contributing features are in fact, obtained in an unsupervised manner. Outside a competition setting, we believe there is tremendous potential for deep learning to be applied in this domain.  

%
\bibliographystyle{abbrv}
\bibliography{references}  
%
%

\end{document}